\documentclass[11pt, a4paper]{article}
\usepackage[utf8]{inputenc}
\usepackage{geometry}
\usepackage{amsmath,amssymb,amsfonts}
\usepackage{algorithm}
\usepackage{algorithmic}
\usepackage{graphicx}
\usepackage[table]{xcolor}
\usepackage{booktabs}
\usepackage{multirow}
\usepackage{float}
\usepackage{tikz}
\usetikzlibrary{positioning,shapes,arrows}
\usepackage{pgfplots}
\pgfplotsset{compat=1.17}
\usepackage{caption}
\usepackage{bm}
\usepackage[colorlinks=true,linkcolor=blue,citecolor=blue,urlcolor=blue]{hyperref}

\definecolor{supervised}{RGB}{65, 105, 225}
\definecolor{language}{RGB}{220, 20, 60}
\definecolor{self}{RGB}{46, 139, 87}
\definecolor{supervisedlight}{RGB}{200, 215, 245}
\definecolor{languagelight}{RGB}{255, 200, 210}
\definecolor{selflight}{RGB}{200, 235, 215}
\definecolor{headercolor}{RGB}{240, 240, 240}
\definecolor{finetuned}{RGB}{210, 150, 152}

\title{\textbf{Person Re-ID in 2025: Supervised, Self-Supervised, and Language-Aligned—What Works?}}

\author{Lakshman Balasubramanian\\
MoiiAi Inc.\\
\texttt{lakshman@moiiai.com}}

\date{\today}

\begin{document}

\maketitle

\begin{abstract}
Person Re-Identification (ReID) remains a challenging problem in computer vision. This work reviews various training paradigm and evaluates the robustness of state-of-the-art ReID models in cross-domain applications and examines the role of foundation models in improving generalization through richer, more transferable visual representations. We compare three training paradigms, supervised, self-supervised, and language-aligned models. Through the study the aim is to answer the following questions: Can supervised models generalize in cross-domain scenarios? How does foundation models like SigLIP2 perform for the ReID tasks? What are the weaknesses of current supervised and foundational models for ReID? We have conducted the analysis across 11 models and 9 datasets. Our results show a clear split: supervised models dominate their training domain but crumble on cross-domain data. Language-aligned models, however, show surprising robustness cross-domain for ReID tasks, even though they are not explicitly trained to do so. Code and data available at: \url{https://github.com/moiiai-tech/object-reid-benchmark}.
\end{abstract}

\section{Introduction}

Person Re-Identification (ReID) is the task of recognizing the same individual across different cameras and at different points in time. While humans perform this task effortlessly, it remains a challenging problem in computer vision~\cite{deeplearnignpersnreid}. This difficulty largely arises from substantial variations in appearance caused by changes in viewpoint, illumination, pose, occlusion, and background clutter. A robust ReID model must be invariant to all these factors while still reliably reidentifying across diverse conditions.

Applications of person Re-Identification span a wide range of applications, including smart cities, security and surveillance systems, and retail analytics. Despite strong performance in controlled settings and benchmarks, ReID models~\cite{remix} often suffer from severe performance degradation when deployed in new environments. For e.g., a model trained on surveillance footage from a university campus may fail to reliably recognize identities when applied to a subway station or an outdoor concert.

In this work, we investigate whether such brittleness is inherent to the ReID task itself or a consequence of current modelling and training paradigms. In particular, we examine whether recent advances in foundation models, especially language-aligned vision encoders, can mitigate these limitations and improve robustness across domains.

Our study is structured around four key questions. First, how do supervised, self-supervised, and language-aligned models compare in the context of person ReID? Second, to what extent do supervised ReID models truly generalize beyond their training distributions, as opposed to exploiting dataset-specific biases? Third, why do language-aligned models sometimes outperform highly specialized supervised models on certain benchmarks? Finally, does increasing model scale—in terms of parameters and capacity—lead to consistent improvements in cross-domain ReID performance?

A total of 11 models across 9 datasets—the most comprehensive comparison of training paradigms to date. The evaluation spans across various models like supervised approaches (CLIP-ReID~\cite{clipreid}, OSNet~\cite{osnet}), self-supervised methods (DINOv2~\cite{dinov2}, Perception Encoders~\cite{perceptionencoder}), and language-aligned architectures (CLIP~\cite{clip}, SigLIP2~\cite{siglip2}). The objective of this study is to comprehend the limitations of state-of-the-art model and to furnish practitioners with unambiguous guidance on the utilisation of models in specific circumstances.

\section{Problem Formulation}

\subsection{Formal Definition}

Person ReID tasks assume that the person detection is done by a prior object detection model that can reliably give crops of objects from an image. Give this ReID task can be formulated as a ranking problem, where a query image of an individual is matched against a large gallery of candidate images. The objective is to rank the gallery such that images corresponding to the same individual as the query appear at the top of the list.

The objective is to train a neural network $f_{\bm{\theta}}: \bm{I} \mapsto \bm{z}$ to map images $\bm{I} \in \mathbb{R}^{H \times W \times C}$ into a feature vector $\bm{z} \in \mathbb{R}^{{d}_\text{emb}}$. The objective is to minimize the distance between two images of the same person in the embedding space, while distance between images of different people should be maximized.

Formally, for a query image $\bm{I}$ and a gallery image $\bm{G}_i$, we calculate a similarity score:
\begin{equation}
    s(\bm{I}, \bm{G}_i) = 1 - d(f_{\bm{\theta}}(\bm{I}), f_{\bm{\theta}}(\bm{G}_i))
\end{equation}

where $d$ is a distance function. For an ideal embedding model, an image $\bm{I}$ will be closer to the $\bm{G}_+$ than any imposter $\bm{G}_-$ in the embedding space $\bm{z}$:
\begin{equation}
    d(f_{\bm{\theta}}(\bm{I}), f_{\bm{\theta}}(\bm{G}_+)) < d(f_{\bm{\theta}}(\bm{I}), f_{\bm{\theta}}(\bm{G}_-))
\end{equation}

\subsection{Evaluation Metrics}

\subsubsection{Mean Average Precision (mAP)}

For each query $\bm{I}$ with $K$ correct matches in the gallery, we compute Average Precision (AP),

\begin{equation}
    \text{AP}(\bm{I}) = \frac{1}{K} \sum_{k=1}^{N} P(k) \cdot \bm{I}(\text{match}_k),
\end{equation}

where $P(k)$ is precision at rank $k$, and $\bm{I}(\text{match}_k)$ indicates whether the $k$-th ranked item is a correct match. The mAP is the mean over all queries $\mathcal{Q}$:

\begin{equation}
    \text{mAP} = \frac{1}{|\mathcal{Q}|} \sum_{q \in \mathcal{Q}} \text{AP}(\bm{I}),
\end{equation}

where $\mathcal{Q}$ denotes the set of query samples, $|\mathcal{Q}|$ is the number of queries,

\subsubsection{Cumulative matching Characteristic (CMC)}

Rank-$k$ accuracy measures the percentage of queries where at least one correct match appears in the top $k$ results:

\begin{equation}
\mathrm{Rank}\text{-}k
= \frac{1}{|\mathcal{Q}|}
\sum_{q \in \mathcal{Q}}
\mathbb{I}\!\left(
\exists\, r \in \mathrm{Top}\text{-}k(q)
\;\text{s.t.}\;
y_r = y_q
\right),
\end{equation}

where $\mathrm{Top}\text{-}k(q)$ represents the set of the top $k$ retrieved results for query $q$ ranked by similarity,
$y_q$ and $y_r$ are the identity labels of the query and retrieved sample, respectively,
$\mathbb{I}(\cdot)$ is the indicator function that equals $1$ if the condition holds and $0$ otherwise.

\section{Overview of ReID Methods}

ReID has evolved significantly from early approaches based on hand-crafted features~\cite{handcrafted} to modern methods grounded in deep learning and metric learning. Recent ReID models~\cite{clipreid,osnet,heInstructReIDMultipurposePerson2023, tranreid} aim to learn discriminative embedding spaces in which samples belonging to the same person are closely clustered, while samples from different people are well separated.

While this objective remains consistent, the strategies used to train such models have changed substantially over time. Advances in network architectures, loss functions, and training paradigms have played a central role in the evolution of ReID approaches, leading to increasingly sophisticated approaches for learning identity-preserving representations.

\subsubsection{ Supervised ReID Training}

Supervised ReID methods rely on fully annotated datasets, in which multiple cropped images of the same individual captured from different camera views are explicitly labelled. Under this setting, ReID is commonly formulated as either a classification problem or a metric learning task, with the objective of learning highly discriminative identity-specific representations.

Recent work has pushed this paradigm toward increasingly strong in-domain performance. The work~\cite{tranreid} proposed TransReID, which leverages Vision Transformers to model long-range dependencies and attend to fine-grained visual cues.

Despite achieving state-of-the-art results on in-data benchmarks, supervised ReID models often exhibit limited generalization beyond the training distribution as shown in~\cite{remix}. In particular, they tend to overfit dataset-specific biases, including camera viewpoints, illumination conditions, and background context. As a result, performance can degrade substantially when models are deployed in novel environments with unseen cameras or lighting conditions, highlighting a fundamental limitation of purely supervised training paradigms.

\subsubsection{Domain Generalization Approaches}

To address the concern of domain-shift and cross-domain performance of supervised models, recent works have explored domain generalization strategies that explicitly increase the diversity of training data. Authors in ReMix~\cite{remix}, augment conventional surveillance datasets with a large single camera person dataset. Although these additional samples lack precise identity annotations and consistent multi-camera correspondences, their high variance in appearance, background, and capture conditions discourages the model from exploiting spurious correlations. Such heterogeneous data encourages the model to focus on identity-relevant visual cues that are more robust across domains.

\subsubsection{Fine-Tuning Foundation Vision Encoders}

Another promising direction is to leverage large-scale foundation vision models pretrained on web-scale, diverse datasets. Rather than training a ReID model from scratch, this approach initializes the model with representations learned from broad visual experience. For instance, CLIP~\cite{clip}, is trained on approximately 400 million image-text pairs and encodes rich semantic concepts, such as clothing attributes and accessories. CLIP-ReID~\cite{clipreid} adapts this pretrained representation to the re-identification task through fine-tuning, transferring general visual and semantic knowledge to the task of person matching.

\section{Model Paradigms Under Study}

Given the wide range of available training paradigms, we evaluate three fundamentally different approaches to ReID. Our goal is to assess their robustness and generalization performance beyond the training distribution.

\subsection{Supervised Learning}

Supervised ReID models are generally trained using a combination of an identity classification loss and metric learning objectives.

\paragraph{Identity Classification Loss}
ReID is first formulated as a multi-class classification problem, where the model predicts the identity label of each input image. Given $C$ identities in the training set, the model outputs a $C$-dimensional probability distribution. Cross-entropy loss is used to penalize incorrect predictions,

\begin{equation}
    \mathcal{L}_{\text{ID}} = -\frac{1}{B}\sum_{i=1}^{B}
    \log
    \frac{\exp(\bm{w}_{y_i}^T f_{\bm{\theta}}(\bm{I}_i) / \tau)}
    {\sum_{j=1}^{C} \exp(\bm{w}_j^T f_{\bm{\theta}}(\bm{I}_i) / \tau)},
\end{equation}

where  $\tau$ is a temperature parameter that scales the logits for better optimization, $\bm{w}_j \in \mathbb{R}^d$ is the learnable weight vector (classifier prototype) associated with the $j$-th identity. It acts as a linear classifier in the feature space. Specifically, $\bm{w}_{y_i}$ is the weight vector corresponding to the true class of sample $i$.$B$ denotes the batch size, $C$ is the number of identities, $f_{\bm{\theta}}(\cdot)$ is the feature extractor parametrized by $\bm{\theta}$, and $\tau$ is a temperature parameter. This objective encourages the model to learn features that are discriminative for each identity in the training set. Such loss is used to train~\cite{osnet}.

\paragraph{Triplet Loss}
While classification loss focuses on separating identities globally, triplet loss directly enforces relative distances in the embedding space. Each triplet consists of an anchor image $\bm{a}$, a positive image $\bm{p}$ of the same identity, and a negative image $\bm{n}$ of a different identity. The objective enforces a margin $\alpha$ between positive and negative pairs:

\begin{equation}
    \mathcal{L}_{\text{triplet}} =
    \sum_{(a,p,n)}
    \max\left(0,
    d(f_{\bm{\theta}}(\bm{a}), f_{\bm{\theta}}(\bm{p}))
    - d(f_{\bm{\theta}}(\bm{a}), f_{\bm{\theta}}(\bm{n}))
    + \alpha \right)
\end{equation}

This loss used in methods like~\cite{clipreid} explicitly shapes the embedding space by pulling samples of the same identity closer together while pushing samples of different identities apart.

The overall supervised training objective is given by:

\begin{equation}
    \mathcal{L}_{\text{supervised}} =
    \mathcal{L}_{\text{ID}} + \lambda \mathcal{L}_{\text{triplet}}
\end{equation}

where $\lambda$ controls the relative importance of the metric learning term. Additional to this methods like~\cite{tranreid} embed the camera view information as additional embeddings to steer the embedding space to be camera aware.

\subsection{Self-Supervised Learning}

Self-supervised methods, such as DINOv2~\cite{dinov2}, learn visual representations without specific identity labels. DINOv2 employs a teacher-student framework, where the student network is trained to match the output distribution of a teacher network across different augmented views of the same image:

\begin{equation}
    \mathcal{L}_{\text{DINO}} =
    -\sum_{\bm{I} \in {B}}
    P_\text{t}(f_{\bm{\theta}}(\bm{I}')) \log P_\text{s}(f_{\bm{\theta}}(\bm{I}'')).
\end{equation}

Here, $P_\text{t}$ and $P_\text{s}$ denote the softmax-normalized outputs of the teacher and student networks, respectively, while $\bm{I}'$ and $\bm{I}''$ represent different augmentations of image $\bm{I}$. The teacher parameters are updated as an exponential moving average of the student parameters, which prevents representational collapse.

Although self-supervised models learn strong general-purpose visual features, they lack explicit identity-level supervision. As a result, they must either be used in a zero-shot setting with similarity-based matching or further fine-tuned for ReID.

\subsection{Language Aligned Vision Encoders}

Vision encoders such as CLIP~\cite{clip} and SigLIP2~\cite{siglip2} learn joint image-text embeddings by training on large-scale image caption datasets. CLIP employs a contrastive loss over batches of image-text pairs,

\begin{equation}
\mathcal{L}_{\text{CLIP}} =
-\frac{1}{2B} \sum_{i=1}^{B}
\left[
\log
\frac{
\exp\!\left(
s\!\left(
f^{I}_{\boldsymbol{\theta}}(\boldsymbol{I}_i),
f^{T}_{\boldsymbol{\theta}}(\boldsymbol{T}_i)
\right)/\tau
\right)
}{
\sum_{j=1}^{B}
\exp\!\left(
s\!\left(
f^{I}_{\boldsymbol{\theta}}(\boldsymbol{I}_i),
f^{T}_{\boldsymbol{\theta}}(\boldsymbol{T}_j)
\right)/\tau
\right)
}
+
\log
\frac{
\exp\!\left(
s\!\left(
f^{I}_{\boldsymbol{\theta}}(\boldsymbol{I}_i),
f^{T}_{\boldsymbol{\theta}}(\boldsymbol{T}_i)
\right)/\tau
\right)
}{
\sum_{j=1}^{B}
\exp\!\left(
s\!\left(
f^{I}_{\boldsymbol{\theta}}(\boldsymbol{I}_j),
f^{T}_{\boldsymbol{\theta}}(\boldsymbol{T}_i)
\right)/\tau
\right)
}
\right].
\end{equation}

SigLIP2 replaces the softmax-based contrastive objective with a sigmoid loss that treats each image-text pair independently,

\begin{equation}
\mathcal{L}_{\text{SigLIP}} =
-\frac{1}{B^2} \sum_{i=1}^{B} \sum_{j=1}^{B}
\left[
y_{ij} \log \sigma
\left(
s\!\left(
f^{I}_{\boldsymbol{\theta}}(\boldsymbol{I}_i),
f^{T}_{\boldsymbol{\theta}}(\boldsymbol{T}_j)
\right)
\right)
+
(1 - y_{ij})
\log
\left(
1 -
\sigma\!\left(
s\!\left(
f^{I}_{\boldsymbol{\theta}}(\boldsymbol{I}_i),
f^{T}_{\boldsymbol{\theta}}(\boldsymbol{T}_j)
\right)
\right)
\right)
\right].
\end{equation}

Here, $\boldsymbol{I}_i$ denotes the $i$-th image in a batch of size $B$, and $\boldsymbol{T}_i$ denotes the corresponding text associated with image $\boldsymbol{I}_i$ (e.g., a caption, prompt, or natural-language description). The functions $f^{I}_{\boldsymbol{\theta}}(\cdot)$ and $f^{T}_{\boldsymbol{\theta}}(\cdot)$ are the image and text encoders parametrized by $\boldsymbol{\theta}$, respectively. The function $s(\cdot,\cdot)$ denotes a similarity measure between image and text embeddings, typically cosine similarity, and $\tau$ is a temperature parameter controlling the sharpness of the similarity distribution. The sigmoid function is denoted by $\sigma(\cdot)$. In the SigLIP loss, $y_{ij} \in \{0,1\}$ indicates whether the image--text pair $(\boldsymbol{I}_i, \boldsymbol{T}_j)$ is a positive (matched) pair.

The underlying hypothesis is that vision-language alignment introduces semantic structure that transfers more effectively across visual domains than purely visual supervision, potentially improving generalization in ReID.

\section{Datasets and Models}
This section discusses the datasets and models used in this paper to analyse the ReID task

\subsection{Dataset}
We evaluate on 9 diverse datasets spanning different domains and difficulty levels.

\subsubsection{Large-Scale Surveillance Datasets}

\paragraph{MSMT17}~\cite{msmt17} Multi-Scene Multi-Time is currently the largest ReID dataset, with 126,441 bounding boxes of 4,101 identities captured by 15 cameras across indoor and outdoor scenes during different times and weather conditions. Images vary significantly in resolution and quality. MSMT17 serves as the training set for our supervised models.

\paragraph{Market-1501}~\cite{market1501} contains 32,668 images of 1,501 identities from 6 cameras at Tsinghua University campus. It has become the standard benchmark for ReID. Images are relatively high quality with clear viewpoints, making it less challenging than MSMT17.

\paragraph{DukeMTMC-reID}~\cite{dukemtmc} is extracted from the DukeMTMC tracking dataset, containing 36,411 images of 1,404 identities from 8 high-resolution cameras at Duke University. It features diverse camera angles and challenging lighting.

\subsubsection{Controlled Environment Datasets}

\paragraph{CUHK03}~\cite{chuk} was collected on the CUHK campus with 14,097 images of 1,467 identities. Uniquely, it provides both manually labelled and automatically detected bounding boxes. We use the detected version, which is more challenging due to imperfect detection.

\paragraph{GRID}~\cite{grid} captures 250 identities in an underground station with 1,275 images from 8 cameras. It is notoriously difficult due to low resolution, poor lighting, and severe occlusion.

\subsubsection{In-the-Wild Datasets}

\paragraph{CelebReID}~\cite{celebritiesreid} contains images of celebrities from red carpet events, street photography, and media sources. Unlike surveillance footage, these are high-resolution, professionally lit images with diverse poses and fashion styles. This dataset tests whether models can handle the shift from security cameras to high-quality photography.

\paragraph{PKU-ReID}~\cite{pkureid} features outdoor campus scenes from Peking University with natural lighting, seasonal variations, and weather changes. It represents a middle ground between controlled surveillance and completely unconstrained scenarios.

\paragraph{LasT}~\cite{last} (Large-scale Attribute SpatioTemporal) emphasizes temporal consistency and attribute-based retrieval across diverse real-world scenarios with varying conditions.

\paragraph{IUSReID}~\cite{lustpersonreid} focuses on extreme challenges: severe occlusion (people partially hidden), drabmic illumination changes, and unusual viewpoints. It represents worst-case deployment scenarios.

\subsection{Models Evaluated}

We evaluate 11 models spanning different paradigms and scales. Table~\ref{tab:models} summarizes their key characteristics.

\begin{table*}[t]
\centering
\caption{Models evaluated in this study, categorized by paradigm and parameter count.}
\label{tab:models}
\scriptsize
\begin{tabular}{llrll}
\toprule
\rowcolor{headercolor}
\textbf{Model} & \textbf{Paradigm} & \textbf{Trainable params.} & \textbf{Architecture} & \textbf{Training Dataset} \\
\midrule

\rowcolor{supervisedlight}
OSNet-x1.0 & Supervised & 2.5M & Omni-Scale Net & Cross-camera \\
\midrule
\rowcolor{finetuned}
CLIP-ReID & Fine-tuned & 87.5M & ViT-B/16 & Cross-camera \\
\midrule
\rowcolor{languagelight}
CLIP-B-32 & Language-Aligned & 151M & ViT-B/32 & Image-text \\
\rowcolor{languagelight}
CLIP-B-16 & Language-Aligned & 150M & ViT-B/16 & Image-text \\
\rowcolor{languagelight}
CLIP-L-14 & Language-Aligned & 428M & ViT-L/14 & Image-text \\
\rowcolor{languagelight}
SigLIP2-256 & Language-Aligned & 375M & ViT-B/16 (256px) & Image-text \\
\rowcolor{languagelight}
SigLIP2-384 & Language-Aligned & 376M & ViT-B/16 (384px) & Image-text \\
\midrule
\rowcolor{selflight}
DINOv2-B14 & Self-Supervised & 87M & ViT-B/14 & Augmented Images \\
\rowcolor{selflight}
DINOv2-L14 & Self-Supervised & 304M & ViT-L/14 & Augmented Images \\
\rowcolor{selflight}
PE-Core-L14 & Self-Supervised & 671M & ViT-L/14 (336px) & Augmented Images \\
\rowcolor{selflight}
PE-Spatial-S16 & Self-Supervised & 22M & ViT-S/16 (512px) & Augmented Images\\
\bottomrule
\end{tabular}
\end{table*}

\subsubsection{Supervised Models}

\paragraph{OSNet} (Omni-Scale Network)~\cite{osnet} is a lightweight architecture (2.5-4.7M) specifically designed for ReID. It uses multiple convolutional streams at different scales to capture both global structure and local details.

\subsubsection{Fine Tuned Foundational Models}

\paragraph{CLIP-ReID}~\cite{clipreid} adapts CLIP's ViT-B/16 architecture for ReID by initializing with CLIP weights, then fine-tuning on ReID datasets with identity classification and triplet losses. With 87.5M parameters, this represents current state-of-the-art supervised transfer learning.

\subsubsection{Language-Aligned Models}

\paragraph{CLIP~\cite{clip}} variants (ViT-B/32, ViT-B/16, ViT-L/14) are evaluated in zero-shot mode using only the image encoder. CLIP was trained on 400M image-text pairs from the internet using contrastive learning. We test three scales: the efficient B/32 (151M parameters), standard B/16 (150M parameters), and large L/14 (428M parameters).

\paragraph{SigLIP2~\cite{siglip2}} improves on CLIP with sigmoid loss instead of softmax, allowing batch-independent learning. We test two input resolutions: 256 (375.2M parameters) and 384 (375.5M parameters).

\subsubsection{Self-Supervised Models}

\paragraph{DINOv2~\cite{dinov2}} represents vision foundation models trained via self-supervised learning. Unlike CLIP, text information is not used here. We evaluate Base (86.6M parameters) and Large (304.4M parameters) variants. These models excel at dense prediction tasks but may lack the semantic structure needed for ReID.

\paragraph{PE-Core-L14-336~\cite{perceptionencoder}}(Perception Encoder Core) is a large-scale self-supervised vision encoder (671.1M parameters) based on ViT-L/14 architecture, designed for robust visual perception tasks. It leverages advanced position encoding and multi-scale feature extraction. 

\paragraph{PE-Spatial-S16-512~\cite{perceptionencoder}}(Perception Encoder Spatial) is a lightweight self-supervised variant (22.0M parameters) based on ViT-S/16 architecture used for specialized spatial tasks like object detection and segmentation.

\section{Experimental Results}
This section discuss the experimental results across all models and datasets
Table~\ref{tab:full_results} presents comprehensive mAP results for all models and datasets. Also finding and analysis from our experiments are also elaborated.

\begin{table*}[t]
\centering
\caption{Mean Average Precision (mAP \%) across all datasets. Bold: best per dataset. Underline: second best. MSMT17 is the training set for supervised models.}
\label{tab:full_results}
\tiny
\setlength{\tabcolsep}{3pt}
\begin{tabular}{lrrrrrrrrrr}
\toprule
\rowcolor{headercolor}
\textbf{Model} & \textbf{Training params} & \textbf{MSMT} & \textbf{Market} & \textbf{Duke} & \textbf{CUHK} & \textbf{GRID} & \textbf{PKU} & \textbf{LasT} & \textbf{IUS} & \textbf{Celeb} \\
\rowcolor{supervisedlight}
OSNet-x1.0 & 2.5M & 3.37 & 83.57 & 19.63 & 5.86 & 21.85 & 1.90 & 2.73 & 0.86 & 3.06 \\
\midrule
\rowcolor{finetuned}
CLIP-ReID & 87.5M & \textbf{66.22} & \underline{50.59} & \textbf{58.28} & \textbf{39.83} & \textbf{38.61} & \textbf{43.68} & \textbf{16.32} & \textbf{13.93} & 7.93 \\
\midrule
\rowcolor{languagelight}
CLIP-B32 & 151M & 0.10 & 0.37 & 0.24 & 0.31 & 2.70 & 1.39 & 0.20 & 0.18 & 0.61 \\
\rowcolor{languagelight}
CLIP-B16 & 150M & 0.11 & 0.43 & 0.26 & 0.29 & 0.98 & 1.30 & 0.28 & 0.19 & 0.68 \\
\rowcolor{languagelight}
CLIP-L14 & 428M & 0.14 & 0.50 & 0.29 & 0.30 & 1.99 & 1.56 & 0.49 & 0.20 & 0.82 \\
\rowcolor{languagelight}
SigLIP2-256 & 375M & 5.64 & 5.67 & 9.11 & 2.81 & 4.56 & 3.52 & 13.69 & 1.46 & \underline{14.23} \\
\rowcolor{languagelight}
SigLIP2-384 & 376M & 4.56 & 5.97 & 8.36 & 2.77 & 5.72 & 2.23 & \underline{14.01} & 1.31 & \textbf{15.32} \\
\midrule
\rowcolor{selflight}
DINOv2-B14 & 87M & 0.37 & 1.71 & 1.03 & 0.32 & 4.69 & 1.20 & 4.12 & 0.63 & 3.68 \\
\rowcolor{selflight}
DINOv2-L14 & 304M & 0.39 & 1.40 & 0.86 & 0.43 & 3.86 & 1.08 & 4.70 & 0.69 & 3.79 \\
\rowcolor{selflight}
PE-Core-L14 & 671M & 0.91 & 1.25 & 0.74 & 0.57 & 2.92 & 1.47 & 3.91 & 0.44 & 2.26 \\
\rowcolor{selflight}
PE-Spatial-S16 & 22M & 0.09 & 0.54 & 0.20 & 0.31 & 1.13 & 0.88 & 0.94 & - & 0.97 \\
\bottomrule
\end{tabular}
\end{table*}

\subsection{Research Question 1: How do supervised, self-supervised, and language-aligned models compare for person ReID?}

The three paradigms show fundamentally different characteristics. The three paradigms exhibit fundamentally different characteristics, and no single paradigm consistently dominates across all settings. The choice of approach depends critically on the deployment scenario and target domain. Supervised, self-supervised, and language-aligned vision encoders each offer distinct advantages but also suffer from inherent limitations when applied to cross-domain scenarios. In particular, methods based on a single paradigm often fail to generalize effectively across domains. Empirical evidence suggests that hybrid approaches—such as CLIP-ReID models that integrate visual and language supervision with fine tuning on cross camera dataset—achieve superior cross-domain performance by leveraging complementary strengths from multiple paradigms.

\paragraph{Finding 1.1: Supervised models fail in cross-domain dataset}

The performance disparity observed for OSNet highlights a pronounced overfitting effect. While OSNet achieves 83.57\% mAP on Market-1501, its performance drops sharply to 3.37\% mAP on MSMT17. Given its relatively small model size (2.5M parameters, approximately $35\times$ fewer than CLIP-ReID), this behaviour suggests that the model has over-specialized to dataset-specific characteristics of Market-1501. Such results underscore the limitations of optimizing solely for benchmark performance, as models tuned in this manner may fail to generalize to real-world deployment scenarios.

\paragraph{Finding 1.2: Stable Cross-Domain Performance of Language-Aligned Models}

Although SigLIP2 does not achieve the best performance on any individual benchmark, it demonstrates consistent and stable behaviour across all evaluated datasets (in-domain and cross-domain). Specifically, SigLIP2 attains mAP values ranging from 2.8\% to 14.2\% across diverse and visually heterogeneous test sets especially in challenging datasets like LasT and CelebReid. This consistency suggests that language-aligned representations offer a robust baseline when the target deployment domain is unknown, trading peak performance for improved reliability under distribution shift.

\paragraph{Finding 1.3: Self-Supervised Models Underperform in Zero-Shot ReID}

Despite strong performance on a wide range of generic vision benchmarks, DINOv2 exhibits limited effectiveness in the zero-shot ReID setting, achieving only 0.3\%-4.7\% mAP across evaluated datasets. This result indicates that purely visual self-supervised pretraining lacks the explicit semantic and relational structure required for reliable identity matching. Notably, DINOv2 performs comparatively better on more challenging datasets, such as GRID (4.69\%) and LasT (4.12\%-4.70\%), suggesting that its representations capture local visual patterns that may be beneficial under adverse conditions, albeit insufficient for high-precision ReID.

\paragraph{Finding 1.4: Hybrid Approaches are the best for ReID tasks currently}
Hybrid approach like CLIP-ReiD fine tuned the language-aligned and visual representations for the ReID tasks, consistently outperforms traditional supervised models such as OSNet-x1.0 and foundational models like Siglip2 across nearly all evaluated benchmarks. CLIP-ReID demonstrates that hybrid models leveraging both visual and language cues provide superior cross-domain generalization, confirming the advantage of combining paradigms for robust person re-identification.

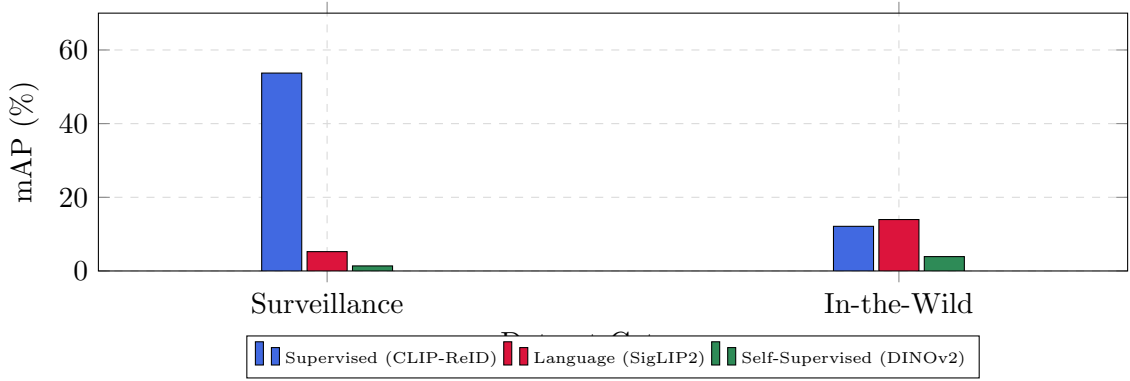
\begin{figure}[H]
\centering
\begin{tikzpicture}
\begin{axis}[
    width=0.95\columnwidth,
    height=5cm,
    xlabel={Dataset Category},
    ylabel={mAP (\%)},
    symbolic x coords={Surveillance, In-the-Wild},
    xtick=data,
    ymin=0, ymax=70,
    legend style={at={(0.5,-0.25)}, anchor=north, legend columns=3, font=\tiny},
    bar width=15pt,
    ybar,
    enlarge x limits=0.4,
    grid=major,
    grid style={dashed, gray!30}
]

\addplot[fill=supervised, draw=black] coordinates {
    (Surveillance, 53.73) (In-the-Wild, 12.12)
};

\addplot[fill=language, draw=black] coordinates {
    (Surveillance, 5.25) (In-the-Wild, 13.96)
};

\addplot[fill=self, draw=black] coordinates {
    (Surveillance, 1.37) (In-the-Wild, 3.90)
};

\legend{Supervised (CLIP-ReID), Language (SigLIP2), Self-Supervised (DINOv2)}
\end{axis}
\end{tikzpicture}
\caption{Cross-domain performance degradation. Supervised models excel on surveillance but catastrophically fail in-the-wild. Language-aligned models show the opposite pattern. Surveillance = avg(MSMT17, Market, Duke, CUHK03). In-the-Wild = avg(LasT, CelebReID).}
\label{fig:domain_shift}
\end{figure}

\begin{figure}[H]
\centering
\begin{tikzpicture}
\begin{axis}[
    width=0.95\columnwidth,
    height=5.5cm,
    xlabel={Target Dataset},
    ylabel={Performance Retention (\%)},
    symbolic x coords={Market, Duke, PKU, CUHK03, GRID, LasT, IUS, Celeb},
    xtick=data,
    x tick label style={rotate=45, anchor=east, font=\scriptsize},
    ymin=0, ymax=270,
    legend style={at={(0.5,1.05)}, anchor=south, legend columns=2, font=\tiny},
    grid=major,
    grid style={dashed, gray!30},
    ylabel style={font=\small}
]

\addplot[color=supervised, mark=*, thick] coordinates {
    (Market, 57.9) (Duke, 69.5) (PKU, 59.1) (CUHK03, 42.4)
    (GRID, 48.3) (LasT, 13.2) (IUS, 15.3) (Celeb, 8.8)
};

\addplot[color=language, mark=square*, thick] coordinates {
    (Market, 100.5) (Duke, 161.5) (PKU, 62.4) (CUHK03, 49.8)
    (GRID, 80.9) (LasT, 242.7) (IUS, 25.9) (Celeb, 252.3)
};

\legend{CLIP-ReID (vs MSMT17), SigLIP2 (vs avg)}
\end{axis}
\end{tikzpicture}
\caption{Cross-domain generalization patterns. CLIP-ReID retention rate calculated as (target\_mAP / MSMT17\_mAP) × 100. Performance drops sharply for out-of-domain datasets (LasT, Celeb). SigLIP2 shows opposite pattern with higher retention on diverse datasets, with values $\geq$ 100\% indicating better performance than baseline.}
\label{fig:generalization}
\end{figure}
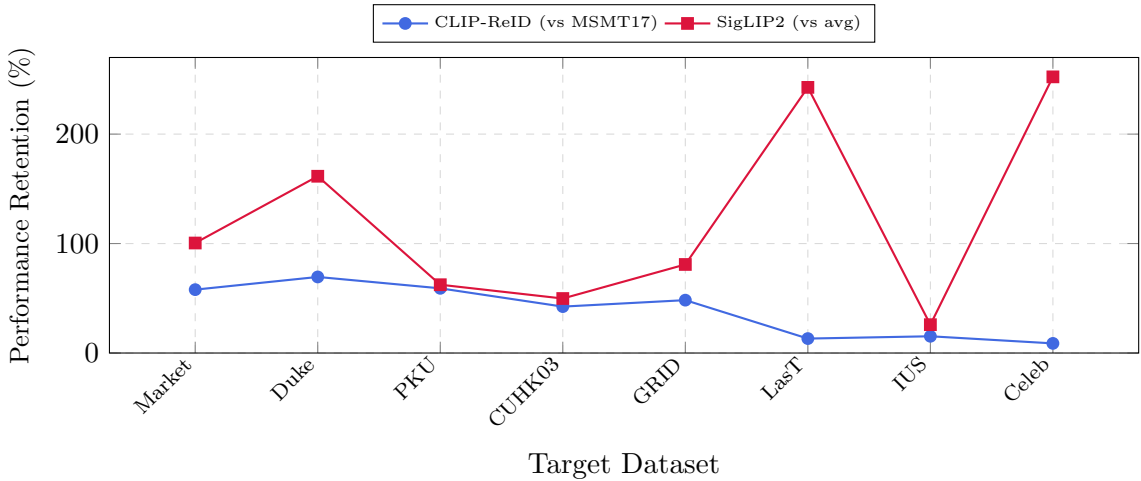

\subsection{Research Question 2: Why Do Language-Aligned Models Outperform Supervised Specialists on Certain Datasets?}

Language-aligned models exhibit unexpected robustness by leveraging semantic understanding rather than relying solely on pixel-level pattern matching. This section analyzes the conditions under which such models outperform fully supervised ReID specialists and provides insight into the underlying mechanisms.

\paragraph{Finding 2.1: Strong Performance on In-the-Wild Datasets}

Despite not being trained explicitly for person re-identification, SigLIP2 substantially outperforms supervised ReID models on several in-the-wild datasets. Notably, SigLIP2 achieves:

\begin{itemize}
    \item CelebReID: 14.23\% mAP compared to 7.93\% for supervised models
\end{itemize}

These results can be attributed to the pre-training of language-aligned models on large-scale web data, which likely includes information about celebrities present in CelebReID. As a result, SigLIP2 already possesses prior knowledge of these identities, enabling it to achieve higher mAP scores compared to other models. This demonstrates that, in scenarios where the target domain overlaps with data seen during large-scale pre-training, language-aligned models can outperform specialized supervised models, even without explicit ReID training.

\paragraph{Finding 2.2: Data Distribution Dominates Task-Specific Supervision}

A key factor underlying this performance gap is the alignment between pretraining data distributions and the target domain. CelebReID consists primarily of high-quality, professionally captured images from known people, including red carpet photography and media portraits. Such imagery closely resembles the web-scale image-text data used to train SigLIP.

In contrast, datasets such as MSMT17 comprise low-resolution surveillance footage with constrained viewpoints and limited visual diversity. Supervised ReID models trained on similar surveillance datasets tend to overfit domain-specific cues, whereas language-aligned models benefit from broader semantic representations that transfer more effectively across domains.

\paragraph{Finding 2.3: Semantic Representation versus Pixel-Level Pattern matching}

Language-aligned models are trained on image-text pairs describing compositional visual concepts (e.g., ``woman wearing a red evening gown'' or ``man in a leather jacket with sunglasses''). As a result, they learn disentangled and transferable semantic attributes, including:

\begin{itemize}
    \item \textit{Personhood}: body structure and pose
    \item \textit{Attributes}: color, texture, and material invariance
    \item \textit{Objects}: clothing categories and accessories
    \item \textit{Relations}: spatial and compositional structure (e.g., ``wearing'')
\end{itemize}

By contrast, supervised ReID models often rely on dataset-specific correlations, such as localized colour patterns or background cues associated with a particular identity. Consequently, language-aligned models match individuals based on high-level semantic appearance rather than brittle, low-level visual patterns, resulting in improved robustness under domain shift.

\paragraph{Finding 2.4: Architectural and Objective Differences between SigLIP and CLIP}

Architectural and training objective choices further influence zero-shot transfer performance. Vanilla CLIP, without task-specific fine-tuning, performs poorly across all evaluated datasets, achieving only 0.1\%-2.7\% mAP. CLIP-ReID, which incorporates ReID-specific fine-tuning, substantially improves performance, highlighting the importance of task adaptation.

However, SigLIP2 consistently outperforms CLIP in the zero-shot setting, achieving 5\%-14\% mAP compared to CLIP's 0.1\%-2\%. This can be attributed to SigLIP2's additional training objectives like dense captioning, mask-level patch representation learning and spatial understanding tasks. This design choice appears to yield more transferable and robust visual representations, particularly in cross-domain ReID scenarios.

\begin{figure}[H]
\centering
\begin{tikzpicture}
\begin{axis}[
    width=\columnwidth,
    height=6cm,
    ybar,
    bar width=12pt,
    xlabel={Model Type},
    ylabel={Best mAP (\%) by Category},
    symbolic x coords={Supervised, Language Aligned, Self-Supervised},
    xtick=data,
    ymin=0, ymax=70,
    legend style={at={(0.5,1.05)}, anchor=south, legend columns=2, font=\tiny},
    grid=major,
    grid style={dashed, gray!30},
    xlabel style={font=\small},
    ylabel style={font=\small}
]

\addplot[fill=supervised!70, draw=black] coordinates {
    (Supervised, 66.02) (Language Aligned, 5.64) (Self-Supervised, 0.91)
};
\addlegendentry{MSMT17 (Training)}

\addplot[fill=language!70, draw=black] coordinates {
    (Supervised, 8.74) (Language Aligned, 14.01) (Self-Supervised, 4.70)
};
\addlegendentry{LasT (In-Wild)}

\addplot[fill=self!70, draw=black] coordinates {
    (Supervised, 5.87) (Language Aligned, 15.32) (Self-Supervised, 3.79)
};
\addlegendentry{CelebReID (In-Wild)}

\end{axis}
\end{tikzpicture}

\caption{Best model performance by paradigm across dataset types. Supervised models dominate on their training domain (MSMT17) but collapse on in-the-wild datasets. Language-aligned models show the opposite pattern, excelling where supervised models fail.}
\label{fig:paradigm_strengths}
\end{figure}
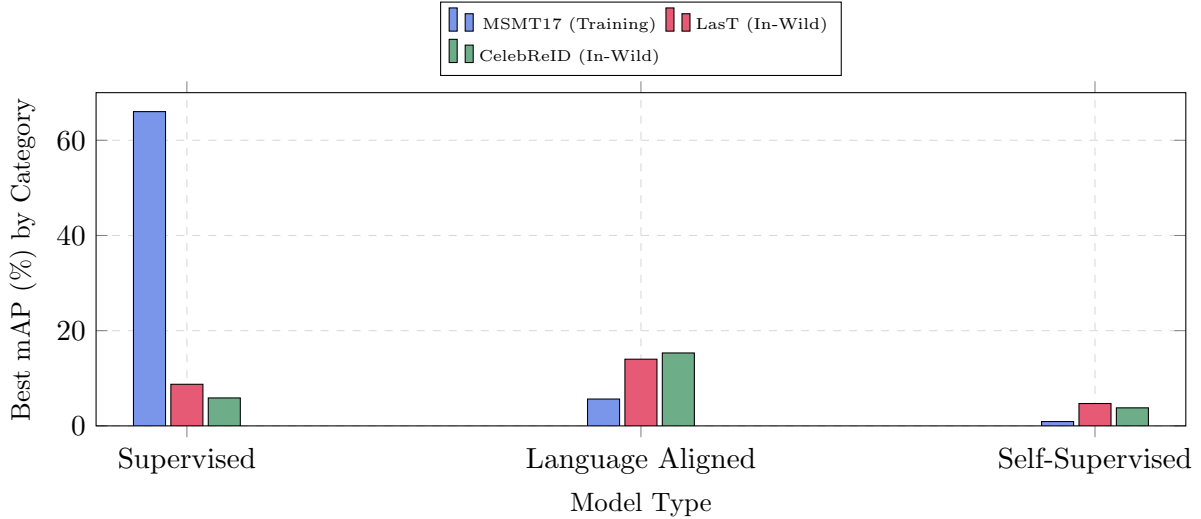

\subsection{Research Question 3: What Role Do Model Size and Architectural Choices Play in Generalization?}

Our results challenge the common assumption that increasing model size necessarily leads to improved generalization. Instead, they indicate that training paradigm and data distribution play a substantially more important role than raw parameter count.

\paragraph{Finding 3.1: Model Scale Does Not Guarantee Improved Performance}

A comparison of model sizes within and across training paradigms reveals clear diminishing returns from increased scale:

\begin{itemize}
    \item OSNet (2.5M parameters) outperforms models that are over two orders of magnitude larger on its target dataset.
    \item CLIP-L/14 (428M parameters) achieves only marginal improvements over CLIP-B/16 (150M parameters), with both models failing to perform effectively in the zero-shot ReID setting.
    \item DINOv2-Large (304M parameters) exhibits performance comparable to DINOv2-Base (87M parameters).
    \item PE-Core (671M parameters), despite being the largest evaluated model, achieves only mediocre performance.
    \item SigLIP2 trained at higher input resolution yields modest gains of approximately 1\% mAP over 256 px, while incurring roughly twice the computational cost.
\end{itemize}

These results suggest that a smaller model trained on data well-aligned with the target distribution can outperform substantially larger models trained on mismatched data.

\paragraph{Finding 4.2: Architectural Complexity Is Secondary to Training Paradigm}

Architectural modifications alone do not consistently improve performance. For instance, increasing model capacity without corresponding improvements in training data or supervision can be counterproductive. These findings suggest that specialized architectural choices require appropriately structured training signals to be effective.

\paragraph{Finding 4.3: Absence of a Unified Model for Performance and Generalization}

A notable performance gap exists between supervised ReID models and language-aligned approaches. Supervised models achieve strong performance on surveillance datasets (e.g., up to 66\% mAP on MSMT17) but generalize poorly to diverse domains. Conversely, language-aligned models demonstrate robust cross-domain behaviour (5\%-14\% mAP across heterogeneous datasets) but underperform on traditional surveillance benchmarks.

Currently, no evaluated model achieves both high performance on surveillance-style datasets and strong generalization across diverse visual domains. This gap highlights an opportunity for hybrid approaches that combine the discriminative strength of supervised ReID training with the semantic robustness conferred by language-aligned representations.

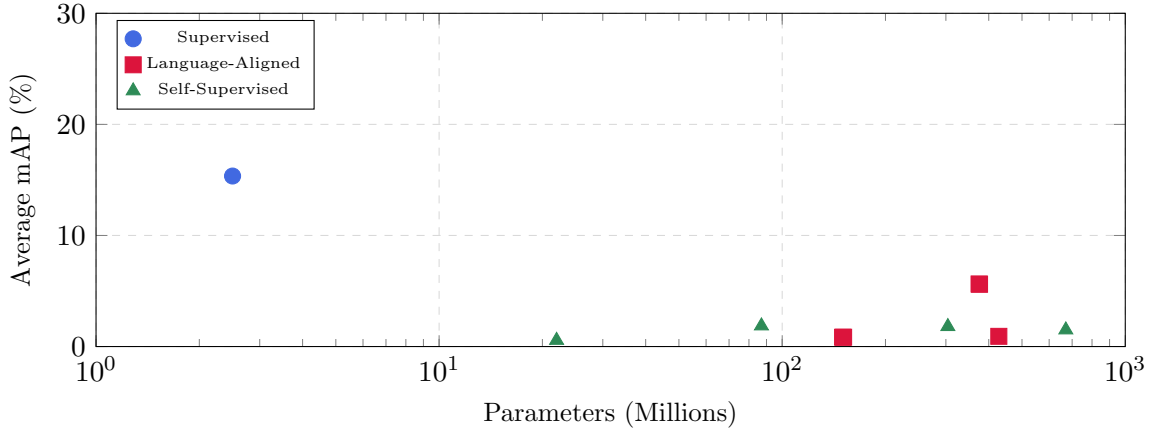
\begin{figure}[H]
\centering
\begin{tikzpicture}
\begin{axis}[
    width=0.95\columnwidth,
    height=6cm,
    xlabel={Parameters (Millions)},
    ylabel={Average mAP (\%)},
    xmode=log,
    log basis x=10,
    xmin=1, xmax=1000,
    ymin=0, ymax=30,
    legend style={at={(0.02,0.98)}, anchor=north west, font=\tiny},
    grid=major,
    grid style={dashed, gray!30},
    xlabel style={font=\small},
    ylabel style={font=\small}
]

\addplot[only marks, mark=*, mark size=3pt, color=supervised] coordinates {
    (2.5, 15.35)   
    (87.5, 37.26)  
};

\addplot[only marks, mark=square*, mark size=3pt, color=language] coordinates {
    (151, 0.84)    
    (150, 0.84)    
    (428, 0.92)    
    (375, 5.63)    
    (376, 5.61)    
};

\addplot[only marks, mark=triangle*, mark size=3pt, color=self] coordinates {
    (87, 1.86)     
    (304, 1.80)    
    (671, 1.50)    
    (22, 0.56)     
};

\legend{Supervised, Language-Aligned, Self-Supervised}
\end{axis}
\end{tikzpicture}
\caption{Model size vs. average performance across all datasets. Larger models do not guarantee better performance. Supervised models (blue circles) achieve highest performance despite modest size. PE-Core (671M, green triangle) underperforms despite being the largest model.}
\label{fig:scale_vs_performance}
\end{figure}

\section{Future Research Directions}

Our results reveal a clear performance gap: supervised models excel on in-domain data but fail on cross-domain data, while language-aligned models provide moderate performance everywhere. This suggests the need for advanced hybrid approaches that effectively combine supervised discriminative power with language-aligned semantic robustness, avoiding the catastrophic forgetting of semantic priors observed in standard fine-tuning methods like CLIP-ReID. Training ReID-specific foundation models on diverse datasets, weakly-supervised web data, and synthetic data could bridge this gap. Moving beyond pure feature extraction, future systems could enable text-guided retrieval ("find person wearing blue shirt"), explainable matching decisions, and temporal reasoning across clothing changes.

Critical challenges remain for real-world deployment: privacy concerns requiring federated learning or encrypted matching, robustness to clothing changes through gait or body shape modeling, long-term tracking across days or weeks, and computational efficiency through distillation and quantization. Addressing these challenges while maintaining the generalization benefits of language-aligned models represents the path forward for practical ReID systems.

\section{Conclusion}

We tested 11 models across 9 datasets to understand how supervised, self-supervised, and language-aligned approaches compare for person re-identification. Here's what we found:

\begin{enumerate}
    \item \textbf{Supervised models dominate their training domain but collapse elsewhere.} CLIP-ReID hits 66\% on MSMT17 and transfers decently (38-46\%) to similar in-domain data. But such methods perform poorly in cross-domain data.

    \item \textbf{Language alignment brings  robustness.} SigLIP2 beats supervised specialists on CelebReID (14.2\% vs 5.8\%) and LasT (13.7\% vs 8.7\%). Semantic understanding from web-scale pretraining generalizes better than domain-specific pattern matching.

    \item \textbf{Self-supervised models do not work for zero-shot ReID.} DINOv2 manages only 0.3-4.7\% mAP. Pure visual self-supervision lacks the discriminative structure needed for identity matching.

    \item \textbf{Bigger is not better.} PE-Core (671M parameters) performs about the same as much smaller models. OSNet (2.5M parameters) wins on its target dataset. Training procedure and data distribution matter more than parameter count.

    \item \textbf{No single paradigm wins everywhere.} Choose based on your deployment: supervised for in-domain, language-aligned for diverse or unknown scenarios.
\end{enumerate}

Future work should focus on hybrid approaches combining supervised discriminative power with language-aligned semantic robustness. 

\section*{Acknowledgments}

This research was conducted at MoiiAi Inc. The author thanks the open-source community for providing model implementations and datasets that made this comprehensive study possible. All code, data, and evaluation scripts are publicly available at \url{https://github.com/moiiai-tech/object-reid-benchmark}.

\bibliographystyle{IEEEtran}
\bibliography{references}

\end{document}